\documentclass[twoside]{article}
\usepackage{aistats2014}

\usepackage{mathrsfs,amsfonts} 
\usepackage{graphicx,amsmath,verbatim,subfigure}
\usepackage{algorithm,algorithmic}
\usepackage[semicolon,authoryear]{natbib}

\def\bw{{\bf {w}}}
\def\bx{{ \bf {x} }}
\def\by{{ \bf {y} }}

\def\Y{\mathscr Y}
\def\X{\mathscr X}
\def\W{\mathscr W}

\def\C{\mathscr C}
\def\I{\mathscr I}

\def\Ostar{{\mathscr O}^*}
\def\Ohat{\hat{\mathscr O}}

\def\dsp{\;\;}

\def\byhat{\hat{ \bf {y} }}

\def\xijstar{\xi_j^*}
\def\byjstar{{ \bf {y} }_j^*}
\def\liyi{\delta_i({\by}_i,{\by})}
\def\liyjstar{\delta_j({\byjstar},{\by})}
\def\delfiyi{\Delta f_i ({\by}_i,{\by})}
\def\delfjyjstar{\Delta f_j ({\byjstar},{\by})}

\def\Clbyl{\frac{C_l}{l}}
\def\Cubyu{\frac{C_u}{u}}

\begin{document}

%

%

\twocolumn[

\aistatstitle{Large Margin Semi-supervised Structured Output Learning}
\aistatsauthor{ Balamurugan P. \And Shirish Shevade \And Sundararajan Sellamanickam }

\aistatsaddress{Indian Institute of Science, India \And Indian Institue of Science, India \And Microsoft Research, India } ]



\begin{abstract}
  In structured output learning, 
  obtaining labeled data for 
  real-world applications is usually costly,  
  while unlabeled examples are  
  available in abundance.     
  Semi-supervised structured classification 
  has been developed to handle large amounts of 
  unlabeled structured data. 
  In this work, we consider semi-supervised structural SVMs with 
  domain constraints. 
  The optimization problem, which in general is not convex, 
  contains the loss terms 
  associated with the labeled and unlabeled examples along with the 
  domain constraints. 
  We propose a simple optimization approach, 
  which alternates between solving a 
  supervised learning problem and a constraint matching problem.  
  Solving the constraint matching problem is difficult for structured prediction, 
  and we propose an 
  efficient and effective hill-climbing method 
  to solve it. 
  The alternating optimization is carried out within a 
  deterministic annealing framework, which helps in 
  effective constraint matching, 
  and avoiding local minima which are not very useful.  
  The algorithm is simple to implement and achieves comparable generalization
  performance on benchmark datasets. 
%
\end{abstract}

\section{Introduction}

Structured classification involves learning a classifier to 
predict objects like trees, graphs and image segments.  
Such objects are usually composed 
of several components with  
complex interactions, and are 
hence called ``structured''. 
Typical structured classification 
techniques learn from 
a set of labeled training examples $\{(\bx_i,\by_i)\}_{i=1}^{l}$,  
where the instances $\bx_i$ are from an input space $\X$ 
and the corresponding labels $\by_i$ belong to 
a structured output space $\Y$. 
Several efficient algorithms 
are available for fully supervised structured 
classification (see for \textit{e.g.}~\cite{joachims_cp, sdm}).
In many practical applications, however, obtaining the label  
of every training example is a tedious task and we are often left with 
only a very few labeled training examples. 

When the training set contains only a few training  
examples with labels, and a large number of 
unlabeled examples, a common 
approach is to use semi-supervised 
learning methods~\citep{chapellebook}. 
For a set of labeled training examples $\{(\bx_i,\by_i)\}_{i=1}^{l}$, 
$\bx_i \in \X$, $\by_i \in \Y$ 
and a set of unlabeled examples $\{\bx_j\}_{j=l+1}^{l+u}$, 
we consider the following semi-supervised 
learning problem:
\begin{align}
\min_{{\bw}, \by_j^* \in \Y} \dsp \frac{1}{2} {\| {\bw} \|^2} + \Clbyl \sum_{i=1}^{l} L_s(\bx_i, \by_i;\bw) \nonumber \\ 
+ \Cubyu \sum_{j=l+1}^{l+u} L_u(\bx_j, \by_j^*;\bw) \nonumber \\ 
{\rm s.t.} \dsp \by_j^* \in \W, \; \forall \; j=l+1,\ldots,l+u, \; \nonumber \\ 
\W = \bigcup_{k} \W_k  \nonumber \\
\label{general_semissvm}
\end{align}
where $L_s(\cdot)$ and $L_u(\cdot)$ 
denote the loss functions corresponding 
to the labeled and unlabeled set of 
examples respectively. 
In addition to minimizing 
the loss functions, we also want to ensure that 
the predictions $\by_j^*$ over the unlabeled data satisfy 
a certain set of constraints: $\W=\bigcup_{k} \W_k$, 
determined using domain knowledge. 
Unlike binary or multi-class classification 
problem, the solution of~\eqref{general_semissvm} 
is hard due to each $\by_j^*$ 
having combinatorial possibilities. 
Further, the constraints play a key role 
in semi-supervised 
structured output learning, as demonstrated 
in~\citep{codl, daso}. \cite{codl} also provide 
a list of constraints (Table 1 in their paper), 
useful in sequence labeling. Each of these 
constraints can be expressed using a function, 
$\Phi: \X \times \Y \rightarrow T$, 
where $T=\{0,1\}$ for hard constraint or 
$T=\mathbb{R}$ for soft constraints. 
For example, the constraint that ``a citation can 
only start with author'' is a hard constraint and the violation 
of this constraint can be denoted as $\Phi_1(\bx_j,\by_j^*)=1$. 
On the other hand. the constraint that ``each output 
at has at least one author'' can be expressed as 
$\Phi_2(\bx_j, \by_j^*) \geq 1$. 
Violation of constraints can be penalized 
by using an appropriate constraint loss function 
$\C(\Phi_k(\bx_j,\byjstar)-c)$ in the 
objective function. 
The domain constraints 
$\Phi(\X,\Y)$ can be further divided into two broad categories, 
namely the \textit{instance level} constraints, which 
are imposed over individual training examples,   
and the \textit{corpus level} constraints, imposed over the entire corpus.  

By extending the binary transductive SVM algorithm proposed in~\citep{joachims_transsvm} 
and constraining the relative frequencies of the output 
symbols, 
~\cite{zien_trans_SSVM} proposed to solve the transductive SVM problem 
for structured outputs. 
Small improvement in performance over purely supervised 
learning was observed, possibly 
because of lack of domain dependent prior knowledge. 
\cite{codl} proposed a constraint-driven 
learning algorithm (CODL) by incorporating 
domain knowledge in the constraints and 
using a perceptron style learning algorithm. 
This approach resulted in high performance learning 
using significantly less training 
data.~\cite{alt_proj_bellare} proposed an alternating 
projection method to optimize an objective function, 
which used auxiliary expectation constraints. 
~\cite{ganchev_PR} proposed a posterior regularization (PR) 
method to optimize a similar objective function. 
\cite{chun_nam_yu} considered transductive 
structural SVMs and used a convex-concave procedure 
to solve the resultant non-convex problem. 

Closely related to our proposed method is the Deterministic Annealing for Structured Output (DASO) 
approach proposed in~\citep{daso}. 
It deals with the combinatorial nature of the label space by using relaxed 
labeling on unlabeled data. 
It was found to perform better than the approaches like CODL and PR. 
However, DASO has not been explored 
for large-margin methods. 
Moreover, dealing with combinatorial 
label space is not straightforward 
for large-margin methods 
and the relaxation idea 
proposed in~\citep{daso} 
cannot be easily extended to 
handle large-margin formulation. 
This paper has 
the following important contributions 
in the context of semi-supervised 
large-margin 
structured output learning. 

\textbf{Contributions:} 
In this paper, we propose an efficient 
algorithm to solve semi-supervised 
structured output learning problem~\eqref{general_semissvm} 
in the large-margin setting. 
Alternating optimization steps (fix $\byjstar$ and solve for $\bw$, 
and then fix $\bw$ and solve for $\byjstar$) are used to solve 
the problem. 
While solving~\eqref{general_semissvm} for $\bw$ 
can be done easily using any known algorithm, 
finding optimal $\byjstar$ for a fixed $\bw$ 
requires combinatorial search. 
We propose an 
efficient and effective 
hill-climbing method 
to solve the combinatorial label switching 
problem. Deterministic Annealing is used 
in conjunction with alternating optimization 
to avoid poor local minima. Numerical experiments 
on two real-world datasets demonstrate that the proposed 
algorithm gives comparable or better 
results with those reported in~\citep{daso} and~\citep{chun_nam_yu}, 
thereby making the proposed algorithm a useful alternative for semi-supervised 
structured output learning. 

The paper is organized as follows. 
The next section discusses related work on semi-supervised 
learning techniques for structured output learning. 
Section~\ref{da_section} 
explains the deterministic annealing solution 
framework for semi-supervised training of structural SVMs with domain constraints. 
The label-switching procedure is elaborated in section~\ref{labelswitch_section}. 
Empirical results on two benchmark datasets 
are presented in section~\ref{experiments_section}. 
Section~\ref{conclusion_section} concludes the paper. 

\section{Related Work}
\label{related_work_section}

A related work to our approach is the 
transductive SVM (TSVM) for multi-class and 
hierarchical classification by~\citep{keerthi_coling}, 
where  
the idea of TSVMs in~\citep{joachims_transsvm} 
was extended 
to  
multi-class problems. 
The main challenge for multi-class problems 
was in designing an efficient procedure 
to handle the combinatorial optimization 
involving the labels $y_j^*$ for unlabeled examples. 
Note that for multi-class problems, $y_j^* \in \{1,2,\ldots,k\}$ 
for some $k\geq3$. 
\cite{keerthi_coling} showed that 
the combinatorial optimization for multi-class 
label switching results in 
an integer program, and 
proposed a transportation 
simplex method to solve it approximately.  
However, the transportation simplex method 
turned out to be in-efficient and 
an efficient label-switching procedure 
was given in~\citep{keerthi_coling}.  
A deterministic annealing 
method  
and  
domain constraints in the form of class-ratios 
were also used in the training.   
We note however that 
a straightforward extension of 
TSVM to structured output learning is 
hindered by the complexity of solving 
the associated label switching problem. 
Extending the 
label switching procedure to structured outputs is 
much more challenging, due to their complex structure  
and the large cardinality of the output space. 

Semi-supervised structural SVMs 
considered in 
~\citep{zien_trans_SSVM} 
avoid the combinatorial optimization 
of the structured output labels, 
and instead consider a working set of 
labels. 
We also note that the combinatorial optimization 
of the label space is avoided 
in the recent work on 
transductive structural SVMs 
by~\citep{chun_nam_yu};   
instead, a working set of cutting planes is maintained.  
The other related work 
DASO~\citep{daso} too 
does not consider 
the combinatorial 
problem of the label space directly; 
rather, it solves a problem of the following form:
\begin{align}
\min_{\bw,a} R(\bw) + {\mathbb{E}}_{a} L(\Y;\X, \bw) + \C({\mathbb{E}}_{a} [\Phi(\X,\Y)] - c)   
\label{daso_problem}
\end{align}
where $a$ denotes a 
distribution $a(\Y)$ 
over the label space 
and $L(\Y;\X, \bw)$ is 
considered to be the log-linear loss. 
Including the 
distribution $a(\Y)$ avoids dealing 
with the original label space $\Y$ and 
side-steps the combinatorial problem over label space. 
Hence, to the best of our knowledge, no prior 
work exists, which tackles directly the combinatorial 
optimization problem involving structured outputs. 

\section{Semi-supervised learning of structural SVMs}
\label{da_section}

We consider the sequence labeling 
problem as a running example 
throughout this paper. 
The sequence labeling problem 
is a well-known structured classification problem, 
where a sequence of entities $\bx=(x^1, x^2, \ldots, x^M)$ 
is labeled using corresponding sequence of labels 
$\by=(y^1, y^2, \ldots, y^M)$. 
The labels $\{y^j\}_{j=1}^{M}$ 
are assumed to be from a fixed alphabet $\Omega$ 
of size $|\Omega|$. 
Consider a structured input-output space pair $(\X,\Y)$. 
Given a training set of labeled examples $\{(\bx_i,\by_i)\}_{i=1}^{l} \; \in (\X \times \Y)$, 
structural SVMs 
learn a parametrized classification rule $h:\X \rightarrow \Y$ of the form $h(\bx; \bw)=\arg\max_{\by} \bw^T f(\bx,\by)$ 
by solving the following convex optimization problem: 
\begin{align}
\min_{{\bw}, \xi_i \geq 0} \dsp\dsp \frac{1}{2} {\| {\bw} \|^2} + C\sum_{i=1}^{l} \xi_i(\bx_i,\by_i,\bw) \nonumber \\
{\rm s.t.} \dsp\dsp \xi_i \geq \liyi -{\bw}^T {\delfiyi}, \nonumber \\
 \forall \;i=1,\ldots,l, \; \forall \; \by \in \Y. \nonumber \\
\label{ssvm_primal}
\end{align}
An input $\bx_i$ is associated with the output $\by_i$ 
of the same length and this association is captured using the 
feature vector $f(\bx_i,\by_i)$. The notation  
$\delfiyi$ in~\eqref{ssvm_primal} 
is given by $f(\bx_i,\by_i) - f(\bx_i,\by)$, the difference between the feature vectors 
corresponding to $\by_i$ and $\by$, respectively.  
The notation 
$\liyi=\delta(\bx_i,\by_i,\by)$ is a suitable loss term. For sequence labeling applications, 
$\liyi$ can be chosen to be the Hamming loss function. 
$C>0$ is a regularization constant. 

With the availability of a set of unlabeled examples $\{\bx_j\}_{j=l+1}^{l+u} \in \X$, 
the semi-supervised learning problem for structural SVMs 
is given by: 
\begin{align}
\min_{{\bw}, \xi_i \geq 0, \xijstar \geq 0} \dsp\dsp  \frac{1}{2} {\| {\bw} \|^2} + \Clbyl \sum_{i=1}^{l} \xi_i(\bx_i,\by_i,\bw) \nonumber \\ + {\Cubyu} \sum_{j=l+1}^{l+u} \xijstar(\bx_j,\by_j^*,\bw) \nonumber   \\
{\rm s.t.} \dsp\dsp \xi_i \geq \liyi -{\bw}^T {\delfiyi} \; \nonumber \\ \forall \;i=1,\cdots,l, \; \forall \; \by \in \Y, \nonumber \\
\xijstar = \min_{\byjstar \in \Y} \max_{\by \in \Y} \; \liyjstar-{\bw}^T {\delfjyjstar} \nonumber \\ \; \forall \;j=l+1,\cdots,l+u. \nonumber \\
\label{zien_s3svm_primal}
\end{align}
The problem~\eqref{zien_s3svm_primal} 
was considered in~\citep{zien_trans_SSVM}, 
where a working-set idea was proposed 
to handle the optimization 
with respect to $\xijstar$.  
However, domain knowledge was not incorporated 
into the problem~\eqref{zien_s3svm_primal}. 
We consider the following non-convex 
problem for semi-supervised learning of structural SVMs, which 
contain the domain constraints: 
\begin{align}
\min_{{\bw}, \xi_i \geq 0, \xijstar \geq 0} \dsp\dsp  \frac{1}{2} {\| {\bw} \|^2} + \Clbyl \sum_{i=1}^{l} \xi_i(\bx_i,\by_i,\bw) + \nonumber \\ 
		    {\Cubyu} \sum_{j=l+1}^{l+u} \xijstar(\bx_j,\by_j^*,\bw) + \C (\Phi(\X,\Y) - c) \nonumber   \\
{\rm s.t.} \dsp\dsp \xi_i \geq \liyi -{\bw}^T {\delfiyi} \nonumber \\
 \forall \;i=1,\cdots,l, \; \forall \; \by \in \Y, \nonumber \\
\xijstar = \min_{\byjstar \in \Y} \max_{\by \in \Y} \; \liyjstar-{\bw}^T {\delfjyjstar} \nonumber \\
 \forall \;j=l+1,\cdots,l+u. \nonumber \\
\label{s3svm_primal}
\end{align}
Note that the problem~\eqref{s3svm_primal} 
is an extension of the semi-supervised learning 
problems associated with binary~\citep{joachims_transsvm} 
and multi-class~\citep{keerthi_coling} 
outputs. 
$C_u$, the regularization constant associated with the unlabeled examples, 
is chosen using an annealing procedure, which gradually increases the influence 
of unlabeled examples in the training~\citep{daso,keerthi_coling}.  

The objective function in~\eqref{s3svm_primal} is non-convex and hence we resort to 
an alternating optimization approach, which is an extension of the procedure given in~\citep{keerthi_coling}  
for semi-supervised multi-class and hierarchical classification. 
We note however that this extension is not easy. This will become clear when we describe the 
constraint matching problem.

A supervised learning problem is solved to obtain an initial model $\bw$ 
before the alternating optimization is performed. 
This supervised learning is done only with the labeled examples, by solving~\eqref{ssvm_primal}. 
%
With an initial estimate of $\bw$ available in hand, 
the alternating optimization procedure starts by 
solving the constraint matching problem with respect to $\xijstar$:
\begin{align}
\min_{\xijstar \geq 0} \dsp\dsp  {\Cubyu} \sum_{j=l+1}^{l+u} \xijstar(\bx_j,\by_j^*,\bw) + \C (\Phi(\X,\Y) - c)  \nonumber  \\ 
{\rm s.t.} \dsp\dsp \xijstar = \min_{\byjstar \in \Y} \max_{\by \in \Y} \; \liyjstar-{\bw}^T {\delfjyjstar} \nonumber \\ 
\forall \;j=l+1,\cdots,l+u. \nonumber \\
\label{constraintmatchprob}
\end{align}
Note that the objective function in the problem~\eqref{constraintmatchprob} is linear in $\xijstar$. 
However, it also contains the penalty term on 
domain constraints $\C (\Phi(\X,\Y) - c)$. 
The constraints involving $\xijstar$ are not simple, and 
finding the quantity 
\begin{align}
\xijstar = \min_{\byjstar \in \Y} \max_{\by \in \Y} \; \liyjstar-{\bw}^T {\delfjyjstar} 
\label{xijstar_eqn} 
\end{align}
becomes difficult when the domain constraints $\Phi(\X,\Y)$ are also considered. 
Hence an efficient procedure needs to be designed to solve~\eqref{constraintmatchprob}. 
In the next section, we propose an efficient label-switching procedure 
to find 
a set of candidate outputs $\{\byjstar\}_{j=l+1}^{l+u}$ 
for the constraint matching problem~\eqref{constraintmatchprob}. 
Once we find $\{\byjstar\}_{j=l+1}^{l+u}$, 
the alternating optimization moves on to solve~\eqref{s3svm_primal} for 
$\bw$, with $\{\byjstar\}_{j=l+1}^{l+u}$ remaining fixed. 
After solving~\eqref{s3svm_primal}, the alternating optimization  
solves~\eqref{constraintmatchprob} again, for a fixed $\bw$.  
This procedure is repeated until some suitable 
stopping criterion is satisfied. 
%
%
%

The alternating optimization procedure described above, is carried out 
within a deterministic annealing framework. 
The regularization 
constant $C_u$ associated with the 
unlabeled examples is slowly varied 
in the annealing step.  
Maintaining a fixed value of $C_u$ is often not useful,  
and using cross-validation to find a suitable value for $C_u$ 
is also not possible. 
Hence, the deterministic annealing framework provides a useful 
way to get a reasonable value of $C_u$. 
In our experiments, $C_u$ was varied over the range 
$\{10^{-4},3\times10^{-4},10^{-3},3\times10^{-3},\ldots,1\}$. 
We describe the alternating optimization  
along with the deterministic annealing procedure in Algorithm~\ref{ssl_algo}.

\begin{algorithm}
	\caption{\textit{A Deterministic Annealing -Alternating Optimization algorithm to solve~\eqref{s3svm_primal}}}
	\label{ssl_algo}
	\begin{algorithmic}[1]
 	\renewcommand{\algorithmicrequire}{\textbf{Input:}}
	\STATE Input labeled examples $\{({\bx}_i, {\by}_i)\}^n_{i=1}$ and 
	 unlabeled examples $\{{\bx}_j\}^{l+u}_{j=l+1}$.
	\STATE Input $C_l$, the regularization constant and $\Omega$, the alphabet for labels.
	\STATE Set maxiter = 1000.
	\STATE Obtain $\bw$ by solving the supervised learning problem in \eqref{ssvm_primal}. 
 	\FOR{$C_u = 10^{-4},3\times10^{-4},10^{-3},3\times10^{-3},\ldots,1$}
	  \STATE iter:=1
	  \REPEAT
	  \STATE Obtain ${\{\by_j^*\}}_{j=l+1}^{l+u}$ for unlabeled examples by solving the constraint matching problem~\eqref{constraintmatchprob} using Algorithm~\ref{labelswitch_algo}. \label{const_match_step}  
	  \STATE Obtain $\bw$ by solving~\eqref{s3svm_primal} with $\by_j^*$ as the actual labels for $\bx_j, \; j=l+1,\ldots,l+u$.
	  \STATE iter:=iter+1
	  \UNTIL{the labeling $\by_j^*$ does not change for unlabeled examples or iter$>$maxiter} 
	\ENDFOR  
	\end{algorithmic} 
        \end{algorithm}

\section{Label switching algorithm to solve~\eqref{constraintmatchprob}}
\label{labelswitch_section}

In this section, we describe an efficient approach to solve the 
constraint matching problem given by~\eqref{constraintmatchprob}. 
We note that finding $\xijstar$ using the relation~\eqref{xijstar_eqn} 
involves solving a complex combinatorial optimization 
problem. 
Hence, a useful heuristic is to fix  
a $\byjstar \in \Y$, and find a corresponding $\by \in \Y$, 
such that the quantity $\liyjstar-{\bw}^T {\delfjyjstar}$ 
is maximized with respect to $\byjstar$. 
An important step is to 
find a good candidate for $\byjstar$, such that the constraint 
violation $\C (\Phi(\X,\Y) - c)$ is minimized.  
We propose to choose $\byjstar$ by an iterative label-switching 
procedure, which we describe in detail below.

Since the constraint matching problem 
(Step~8 in Algorithm~\ref{ssl_algo})  
always follows 
a supervised learning step, we have an estimate of $\bw$,  
before we solve the constraint matching problem. 
With this current estimate of $\bw$, we can find 
an initial candidate $\byjstar$ for an unlabeled example $\bx_j$ as: 
\begin{align}
 \byjstar = \arg\max_{\by \in \Y} {\bw}^T f(\bx_j,\by). \label{byjstar_initial}
\end{align}
The availability of candidate $\byjstar$ for all unlabeled examples $j=l+1,\ldots,l+u$, 
makes it possible 
to compute the objective 
term in~\eqref{constraintmatchprob}, 
where $\xijstar$ for the fixed $\byjstar$ 
is obtained using 
\begin{align}
 \xijstar = \arg\max_{\by \in \Y} \liyjstar - {\bw}^T {\delfjyjstar}. \label{xijstar_fixedyjstar}
\end{align}
%
Let us denote the objective value in~\eqref{constraintmatchprob} by $\Ostar$ 
and let the length of the sequence $\bx_j=(x_j^1,x_j^2,\ldots,x_j^M)$ 
be $M$. 
We now iteratively pass over the components $x_j^m$, 
$\forall \; m=1,2,\ldots,M$, 
and switch the label components $y_j^m$. 
Recall that the label components $y_j^m$ 
are from a finite alphabet $\Omega$ of size $|\Omega|$. 
The switching is done randomly by replacing the label component 
$y_j^m$ with a new label $y^r \in \Omega$. 
With this replacement, we compute the constraint 
violation $\C (\Phi(\X,\Y) - c)$ and 
the slack term $\xijstar$. If the replacement causes 
a decrease in the objective value $\Ostar$, 
then we keep the new label $y^r$ for the $m$-th component 
and move on to the next component. 
If there is no decrease in the objective value, 
we ignore the replacement and keep the original label. 
Note that the constraint matching is handled for each replacement 
as follows. 
Whenever a new label is considered for the $m$-th component, 
instance level constraint violation 
can be checked with the new label in a straightforward way. 
Any corpus level constraint violation 
is usually decomposable over the instances 
and can also be handled in a simple way. 
Hence the constraint violation term 
$\C (\Phi(\X,\Y) - c)$ 
can be computed for each replacement. 
This ensures that by switching the labels, we do not violate the constraints 
too much.   
Apart from choosing the label $y^r$ for the $m$-th component randomly, 
the component $m$ itself was randomly selected in our implementation. 
The switching procedure is stopped when 
there is no sufficient decrease in the objective value term in~\eqref{constraintmatchprob} 
or when a prescribed upper limit on the number of label switches is exceeded. 
The overall procedure is illustrated in Algorithm~\ref{labelswitch_algo}. 
  
\begin{algorithm}
	\caption{\textit{A label switching algorithm to solve constraint matching problem~\eqref{constraintmatchprob}}}
	\label{labelswitch_algo}
	\begin{algorithmic}[1]
 	\renewcommand{\algorithmicrequire}{\textbf{Input:}}
	\STATE Input unlabeled example $\bx_j$
	\STATE Input $\bw$, $C_u$
	\STATE Set maxswitches = 1000, numswitches=0
	\FOR{$j=l+1,\ldots,l+u$} 
	  \STATE Find initial candidate $\byjstar$ by~\eqref{byjstar_initial}
	  \STATE Compute slack $\xijstar$ and constraint violation $\C (\Phi(\X,\Y) - c)$.
	\ENDFOR
	\STATE Calculate objective value 
	\begin{align}
	\Ostar = {\Cubyu} \sum_{j=l+1}^{l+u} \xijstar(\bx_j,\by_j^*,\bw) + \C (\Phi(\X,\Y) - c). \nonumber
	\end{align}
	\FOR{$j=l+1,\ldots,l+u$} 
	    \STATE $\byhat_j = \by_j^*$, $M = length(\byhat)$, $mincost = \Ostar$  
 		\FOR{$m=1,\dots,M$}
 		  \STATE $y^m$$=$$m^{th}$ label component of $\byhat$, $mincostlabel$$=$$y^m$
 		  \FOR{$y^r \in \Omega$ and $y^r \neq y^m$}
 		    \STATE Replace $y^m$ with $y^r$.
 		    \STATE Compute $\C(\phi(\X,\Y)$$-$$c)$, the constraint violation.
  		    \STATE Find violator for ${\byhat}_j$ as 
		    \begin{align}
		    {\bar{\by}}=\arg\max_{\by} \{ \bw^T f(\bx_j,\by) + \delta_j(\byhat_j,\by) \}. \nonumber
		    \end{align}
 		    \STATE Compute slack 
		    \begin{align}  
		    \xi(\byhat_j) = \max(0, \bw^T \Delta f_j(\byhat_j, \bar{\by}) + \delta_j(\byhat_j,\bar{\by})). \nonumber
		    \end{align}  
 		    \STATE Compute objective value of~\eqref{constraintmatchprob} as $\Ohat$.
 		    \IF {$\Ohat$$<$$mincost$}
 			\STATE $mincost$ = $\Ohat$, \; $mincostlabel = y^r$
 		    \ENDIF
 		  \ENDFOR
 		  \STATE Replace $m^{th}$ label of $\by_j^*$ with $mincostlabel$ 
 		  \STATE numswitches = numswitches + 1
 		  \IF{numswitches $>$ maxswitches}
		    \STATE Goto Step 30
		  \ENDIF
		\ENDFOR
	\ENDFOR
 	\STATE Output ${\{\byjstar\}}_{j=l+1}^{l+u}$.
	\end{algorithmic} 
        \end{algorithm}

\section{Experiments and Results} 
\label{experiments_section}

We performed experiments with the proposed semi-supervised 
structured classification 
algorithm 
on two benchmark sequence labeling datasets; 
the citations and apartment advertisements.  
These datasets and were originally introduced in~\citep{field_segmentation} 
and contain manually-labeled 
training and test examples. 
The datasets also contain a 
set of unlabeled examples, which  
is proportionally large when compared to 
the training set. 
The annealing schedule was performed as follows: 
the annealing temperature was started at $10^{-4}$ and 
increased in small steps 
and stopped at $1$. 
The evaluation is done in terms of 
labeling accuracy on the test data 
obtained by the model at the end of training. 
We used the sequential dual method (SDM)~\citep{sdm} 
for supervised learning (Step 4 in Algorithm~\ref{ssl_algo}) 
and compared the following methods for our experiments:
\begin{itemize}
 \item Semi-supervised structural SVMs proposed in this paper (referred to as SSVM-SDM)
 \item Constraint-driven Learning~\citep{codl} (referred to as CODL)
 \item Deterministic Annealing for semi-supervised structured classification~\citep{daso} (referred to as DASO) 
 \item Posterior-Regularization~\citep{ganchev_PR} (referred to as PR)
 \item Transductive structural SVMs~\citep{chun_nam_yu} (referred to as Trans-SSVM)
\end{itemize}


The apartments dataset contains 300 sequences 
from craigslist.org. 
These sequences are labeled using a 
set of 12 labels like features, rent, contact, photos, size and restriction. 
The average sequence length is 119. 
The citations dataset contains 500 sequences, 
which are citations of computer science papers. 
The labeling is done from a set of 13 labels  
like author, title, publisher, pages, and journal. 
The average sequence length for citation dataset is 
35. 
%
%
%

The description and split sizes of the datasets  
are given in Table~\ref{datatable}. 
The partitions of citation data was taken 
to be the same as considered in~\cite{daso}. 
For apartments dataset, we considered 
datasets of 5, 20 and 100 labeled examples. 
We generated 5 random partitions for each case 
and provide the averaged results over these partitions. 

\begin{table}[!ht]
    \caption{\textbf{Dataset Characteristics}}
    \label{datatable}
    \begin{center}
    \begin{tabular}{|l|c|c|c|c|}
    \hline %
              Dataset & $n_{labeled}$ & $n_{dev}$ & $n_{unlabeled}$ & $n_{test}$ \\\hline
	    citation & 5; 20; 300 & 100 & 1000  & 100 \\\hline
	    apartments & 5; 20; 100 & 100 & 1000  & 100 \\\hline
    \end{tabular}
    \end{center}
    \end{table}

    \begin{table*}[t]
    \caption{\textbf{Comparison of average test accuracy(\%) obtained from SSVM-SDM with results in \citep{daso} (denoted by $^{\$}$) and \citep{chun_nam_yu} (denoted by $^{*}$) for Citation Dataset.} 
    (I) denotes inductive setting, in which test examples 
    were not used as unlabeled examples for training. (no I) denotes the setting where test examples were used as unlabeled examples for training. 
		      $^{*}$Note that different set of features were considered in~\citep{chun_nam_yu}.}
    \label{citations_comparison_table}
    \begin{center}
    \begin{tabular}{|l|l|l|l|l|l|l|l|}
    \hline %
             $n_{labeled}$ & Baseline CRF$^{\$}$ & Baseline SDM  & DASO$^{\$}$ & SSVM-SDM & PR$^{\$}$ & CODL$^{\$}$ & Trans-SSVM$^{*}$   \\
			   &			&		&	(I)& (I)	 & (I)	& (I)	& (no I) \\\hline\hline
	    5         & 63.1  & 66.82  & \textbf{75.2} & 74.74 & 62.7 & 71 & 72.8     \\\hline
	    20         & 79.1  & 78.25  & 84.9 & \textbf{86.2}  & 76 & 79.4 & 81.4   \\\hline
	    300         & 89.9  & 91.54  & 91.1 & \textbf{92.92} & 87.29 & 88.8 & 92.8      \\\hline
    \end{tabular}
    \end{center}
    \end{table*}
    
    \begin{table*}[t]
     \caption{\textbf{Comparison of average test accuracy(\%) obtained from SSVM-SDM with results in \citep{daso} (denoted by $^{\$}$) and \citep{chun_nam_yu} (denoted by $^{*}$) for Apartments Dataset.} 
    (I) denotes inductive setting, in which test examples 
    were not used as unlabeled examples for training. (no I) denotes the setting where test examples were used as unlabeled examples for training. 
		      $^{*}$Note that different set of features and split sizes were considered in~\citep{chun_nam_yu}.}
    \label{apartments_comparison_table}
    \begin{center}
    \begin{tabular}{|l|l|l|l|l|l|l|l|}
    \hline %
             $n_{labeled}$ & Baseline CRF$^{\$}$ & Baseline SDM  & DASO$^{\$}$& SSVM-SDM & PR$^{\$}$ & CODL$^{\$}$ & Trans-SSVM$^{*}$    \\ 
             &			&		&	(I) &		(I) &	(I) & (I)	& (no I) \\\hline\hline
	    5         & 65.1  & 64.06  & 67.9 & \textbf{68.28} & 66.5 & 66 & Not Available    \\\hline
	    20         & 72.7  & 73.63  & 76.2 & \textbf{76.37}  & 74.9 & 74.6 & Not Available  \\\hline
	    100         & 76.4  & 79.95  & 80 & \textbf{81.93} & 79 & 78.6 & 78.6     \\\hline
    \end{tabular}
    \end{center}
    \end{table*}
\subsection{Description of the constraints}
We describe the instance level and corpus level constraints 
considered for citations data. A similar description 
holds for those of apartment dataset. We used the same set of constraints 
given in~\citep{codl, daso}. 
The constraints considered are 
of the form 
$\Phi(\X,\Y)-c$,
which are further sub-divided into \textit{instance level} 
constraints of the form
\begin{align}
\Phi_I(\X,\Y) = \phi_I(\bx_j,\by_j^*), \; \; j=l+1,\cdots,l+u  
\end{align}
and the \textit{corpus level} constraints of the form  
$\Phi_D(\X,\Y)-c_D$ where 
\begin{align}
\Phi_D(\X,\Y) = \sum_{j=l+1}^{l+u}\phi_D(\bx_j,\by_j^*). 
\end{align}
We consider the following examples for instance level domain constraints. 

1. \textit{AUTHOR label list can only appear at most once in each citation sequence} :
For this instance level constraint, we could consider 
\begin{align}
\phi_{I_1}(\bx_j,\by_j^*) = \text{Number of $AUTHOR$ label lists in $\by_j^*$}  \nonumber
 \end{align}
and the corresponding $c_I$ to be 1. Hence the instance level domain constraint is 
of the form $\phi_{I_1}(\bx_j,\by_j^*) \leq 1$. The penalty function could then be defined as 
 \begin{align}
 \C(\phi_{I_1}(\bx_j,\by_j^*) - 1) = r | \phi_{I_1}(\bx_j,\by_j^*) - 1 |^2
 		   \nonumber
 \end{align}
where $r$ is a suitable penalty scaling factor. We used $r=1000$ 
for our experiments. 

2. \textit{The word CA is LOCATION} :
For this instance level constraint, we could consider 
\begin{align}
\phi_{I_2}(\bx_j,\by_j^*) = \I\text{(Label for word CA in $\by_j^*$} \nonumber \\		
				  \text{== LOCATION)}  \nonumber
 \end{align}
where $\I(z)$ is the indicator function which is 1 if $z$ is true and $0$ otherwise. 
The corresponding $c_I$ for this constraint is set to 1. Hence the instance level domain constraint is 
of the form $\phi_{I_2}(\bx_j,\by_j^*) = 1$. The penalty function could then be defined as 
 \begin{align}
 \C(\phi_{I_2}(\bx_j,\by_j^*) - 1) = r \nonumber
 \end{align}

3. \textit{Each label must be a consecutive list of words and can occur atmost only once} :
For this instance level constraint, we could consider 
\begin{align}
\phi_{I_3}(\bx_j,\by_j^*) = \text{Number of labels which appear} \nonumber \\ \text{more than once as disjoint lists in $\by_j^*$}  \nonumber 
 \end{align}
The corresponding $c_I$ for this constraint is set to 0. Hence the instance level domain constraint is 
of the form $\phi_{I_3}(\bx_j,\by_j^*) = 0$. The penalty function could then be defined as 
 \begin{align}
 \C(\phi_{I_3}(\bx_j,\by_j^*) ) = r |\phi_{I_3}(\bx_j,\by_j^*)|^2 \nonumber
 \end{align}

Next, we consider some corpus level constraints. 

1. \textit{30\% of tokens should be labeled AUTHOR} :
For this corpus level constraint, we could consider 
\begin{align}
\Phi_{D_1}(\X,\Y) = \text{Percentage of $AUTHOR$ labels in $\Y$} \nonumber 
 \end{align}
and the corresponding $c_I$ to be 30. Hence the corpus level domain constraint is 
of the form $\phi_{D_1}(\X,\Y) = 30$. The penalty function could then be defined as 
 \begin{align}
 \C(\phi_{D_1}(\X,\Y) -30 ) = r | \phi_{D_1}(\X,\Y) - 30 |^2 \nonumber
 \end{align}

2. \textit{Fraction of label transitions that occur on non-punctuation characters is 0.01} :
For this corpus level constraint, we could consider 
\begin{align}
\phi_{D_2}(\X,\Y) = \text{Fraction of label transitions} \nonumber \\ \text{that occur on non-punctuation characters}  \nonumber 
 \end{align}
The corresponding $c_I$ for this constraint is set to 0.01. Hence the corpus level domain constraint is 
of the form $\phi_{D_2}(\X,\Y) = 0.01$. The penalty function could then be defined as 
 \begin{align}
 \C(\phi_{D_2}(\X,\Y) - 0.01 ) = r | \phi_{D_2}(\X,\Y) - 0.01 |. \nonumber
 \end{align}

\subsection{Experiments on the citation data}

We considered the citation dataset with 5, 20 and 300 labeled examples, 
along with 1000 unlabeled examples and measured the performance on a test set of 100 examples. 
The parameter $C_l$ was tuned using a development dataset of 100 examples. 
The average performance on the test set was computed by training on five different partitions 
for each case of 5, 20 and 300 labeled examples. 
The average test set accuracy comparison is presented in Table~\ref{citations_comparison_table}.

The results for CODL, DASO and PR are quoted from 
the Inductive setting in~\citep{daso}, 
as the same set of features and constraints in~\citep{daso} are used 
for our experiments  
and test examples were not considered for 
our training.  
With respect to Trans-SSVMs, 
we quote the results for non-Inductive setting from~\citep{chun_nam_yu}, 
in which constraints were not used for prediction. 
However, we have the following important 
differences with Trans-SSVM in terms of 
the features and constraints.  
The feature set used for Trans-SSVM is not 
the same as that used for our experiments.  
Test examples were used for training Trans-SSVMs, 
which is not done for SSVM-SDM.  
Hence, the comparison results in Table~\ref{citations_comparison_table} 
for Trans-SSVM are only indicative. 
From the results in Table~\ref{citations_comparison_table}, 
we see that, for citations dataset with 5 labeled examples, 
the performance of SSVM-SDM is slightly worse when compared to that 
obtained for DASO. However, 
for other datasets, SSVM-SDM achieves 
a comparable performance. 

\begin{figure}[ht]
  \subfigure{
    \begin{minipage}[b]{0.45\linewidth}
      \includegraphics[scale=0.24]{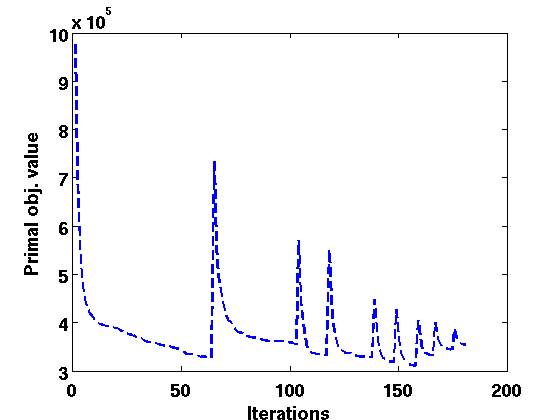}
    \end{minipage}
    \begin{minipage}[b]{0.4\linewidth}
      \includegraphics[scale=0.24]{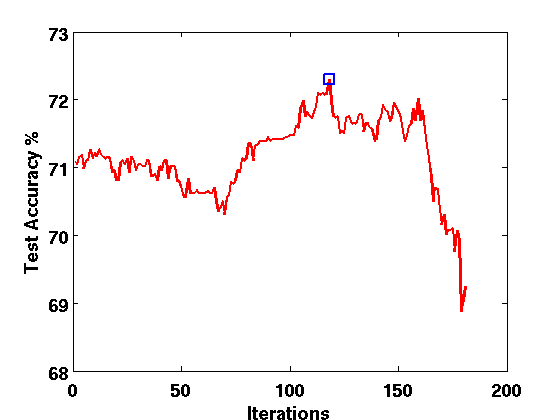}
    \end{minipage}
  } 
  \subfigure{
    \begin{minipage}[b]{0.45\linewidth}
      \includegraphics[scale=0.24]{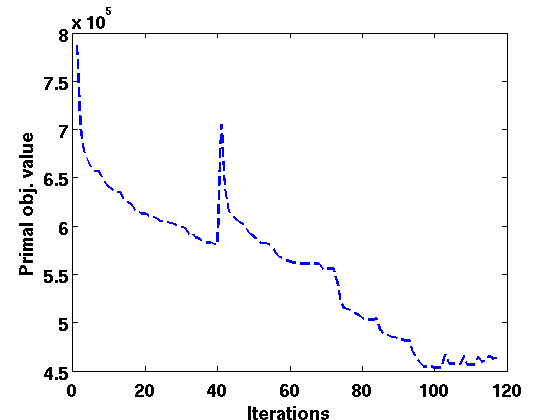}
     \end{minipage}
    \begin{minipage}[b]{0.4\linewidth}
      \includegraphics[scale=0.24]{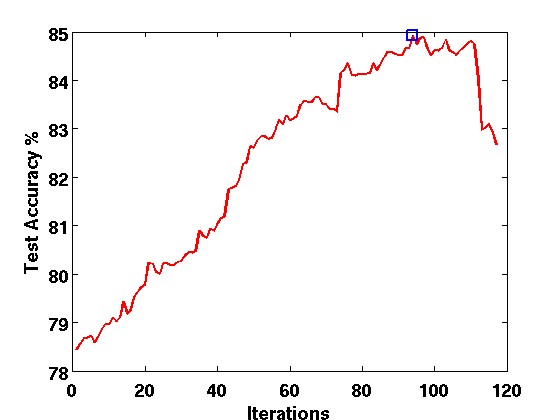}
    \end{minipage}
  }
  \subfigure{
    \begin{minipage}[b]{0.45\linewidth}
      \includegraphics[scale=0.24]{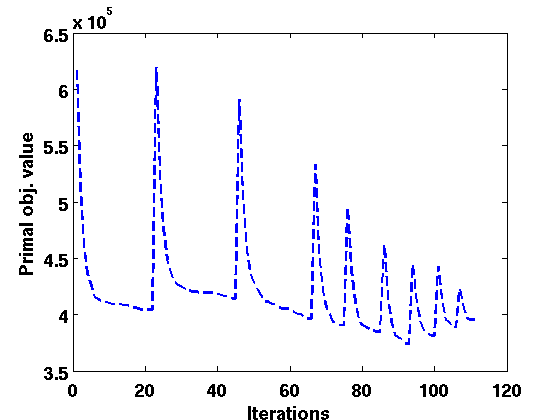}
    \end{minipage}
    \begin{minipage}[b]{0.4\linewidth}
      \includegraphics[scale=0.24]{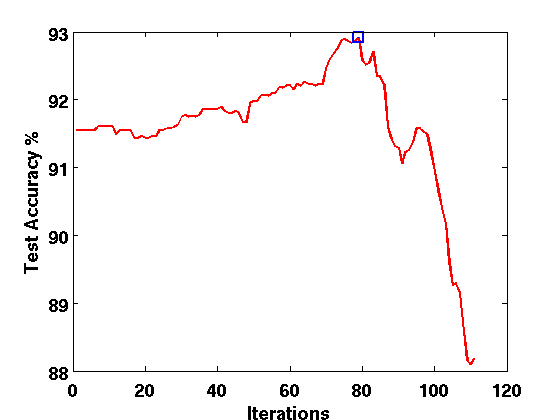}
    \end{minipage}
  }
  \caption{\textbf{Primal objective value and Test accuracy behaviour for a partition of citations dataset.} The rows correspond to 5, 20 and 300 labeled examples in that order. The square marker 
   in the test accuracy plots denotes the best generalization performance.}
  \label{citation_plots}
\end{figure}

We present the plots on test accuracy and primal objective value for the partitions 
containing 5, 20 and 300 examples, in Figure~\ref{citation_plots}.
These plots indicate that as the annealing temperature increases, the generalization 
performance increases initially and 
then continues to drop. 
This drop in generalization performance 
might possibly be the result of over-fitting 
caused by an inappropriate weight $C_u$ for unlabeled examples. 
Similar observation has been made in 
other semi-supervised structured output learning work using 
deterministic annealing~\citep{kai_wei_chang}. 
These observations suggest that finding a suitable stopping criterion 
for semi-supervised structured output learning in the 
deterministic annealing framework 
requires 
further study. 
For our comparison results, we considered the maximum test accuracy obtained from 
the experiments. This is indicated by a square marker in the 
test accuracy plots in Figure~\ref{citation_plots}.

\begin{figure}[ht]
  \subfigure{
    \begin{minipage}[b]{0.45\linewidth}
      \includegraphics[scale=0.24]{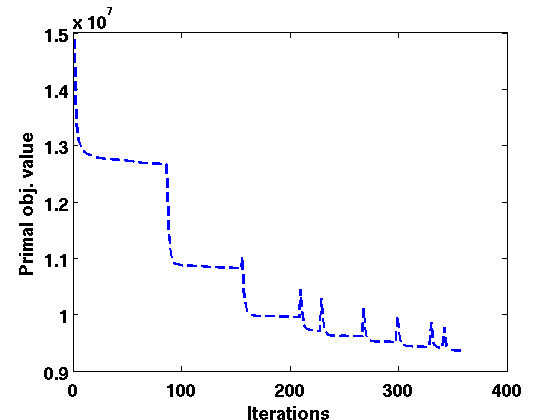}
    \end{minipage}
    \begin{minipage}[b]{0.4\linewidth}
      \includegraphics[scale=0.24]{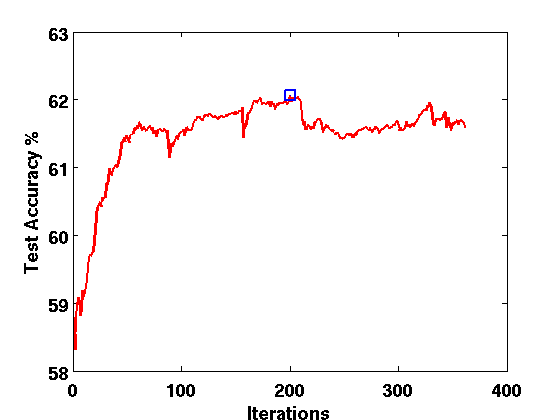}
    \end{minipage}
  } 
  \subfigure{
    \begin{minipage}[b]{0.45\linewidth}
      \includegraphics[scale=0.24]{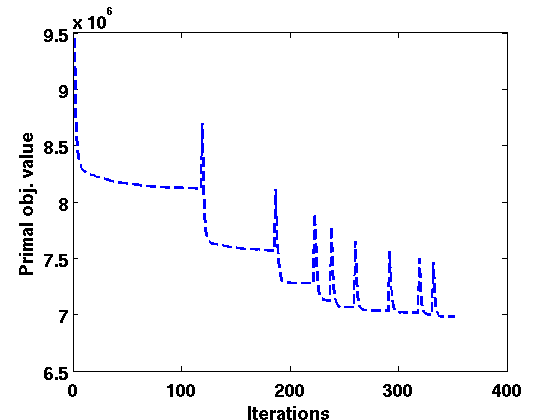}
     \end{minipage}
    \begin{minipage}[b]{0.4\linewidth}
      \includegraphics[scale=0.24]{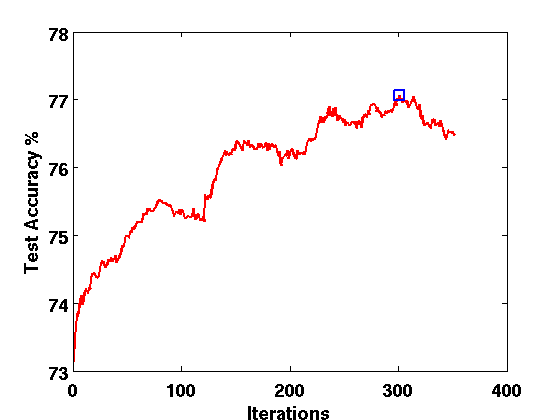}
    \end{minipage}
  }
  \subfigure{
    \begin{minipage}[b]{0.45\linewidth}
      \includegraphics[scale=0.24]{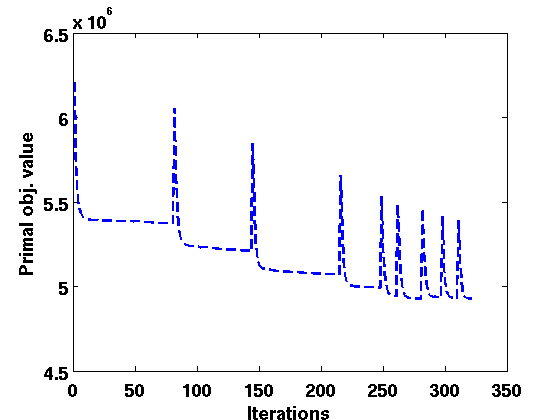}
    \end{minipage}
    \begin{minipage}[b]{0.4\linewidth}
      \includegraphics[scale=0.24]{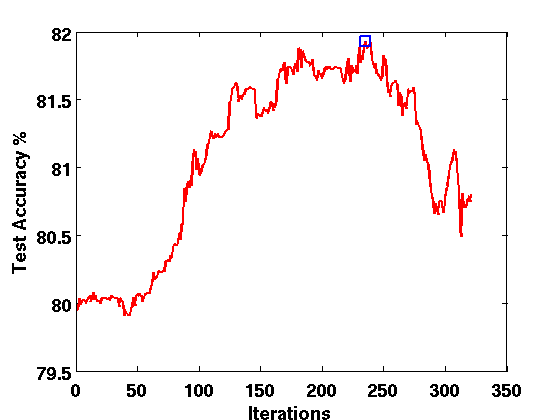}
    \end{minipage}
  }
  \caption{\textbf{Primal objective value and Test accuracy behaviour for a partition of apartments dataset.} The rows correspond to 5, 20 and 100 labeled examples in that order.The square marker 
   in the test accuracy plots denotes the best generalization performance.}
  \label{apartments_plots}
\end{figure}

\subsection{Experiments on the apartments data}

Experiments were performed on the apartments dataset with five partitions each for 5,20 and 100 labeled examples.  
1000 unlabeled examples were considered and a test set of 100 examples was used to measure the generalization performance.  
The parameter $C_l$ was tuned using a development dataset of 100 examples. 
The average test set accuracy comparison is presented in Table~\ref{apartments_comparison_table}.
For apartments dataset, 
though the features and constraints used in our experiments were 
the same as those considered in~\citep{daso}, 
our data partitions differ from those used in their paper.
However, the comparison of mean test accuracy over the 5 different partitions 
for various split sizes is 
justified. 
Note also that we do not include 
the results with respect to Trans-SSVM for some of our experiments, 
as different split-sizes are considered 
for Trans-SSVM in~\citep{chun_nam_yu}. 
In particular,~\cite{chun_nam_yu} considered splits of 10, 25 and 100 labeled examples 
for their experiments. 

The results in Table~\ref{apartments_comparison_table} 
show that SSVM-SDM achieves 
a comparable average performance with  
DASO on all datasets. 
The plots on test accuracy and primal objective value for various partition sizes are given in Figure~\ref{apartments_plots}. 
The plots show a similar performance as seen for the citation datasets.  

\section{Conclusion}
\label{conclusion_section}


In this paper, we considered 
semi-supervised structural SVMs and proposed 
a simple and efficient algorithm 
to solve the resulting optimization problem. 
This involves solving two sub-problems alternately. 
One of the sub-problems is a simple supervised 
learning, performed by fixing the labels 
of the unlabeled training examples. 
The other sub-problem is the constraint matching 
problem in which suitable labeling for unlabeled 
examples are 
obtained. 
This was done by an efficient 
and effective hill-climbing 
procedure, which 
ensures that most of the domain constraints 
are satisfied. 
The alternating optimization was 
coupled with deterministic annealing 
to avoid poor local minima. 
The proposed 
algorithm is  
easy to implement and 
gives comparable generalization performance. 
Experimental results 
on real-world datasets 
demonstrated that the proposed 
algorithm is a useful alternative 
for semi-supervised 
structured output learning. 
The proposed label-switching method 
can also be used to 
handle complex constraints, which 
are imposed over only parts of the structured output. 
We are currently investigating this extension. 

\bibliographystyle{apa}

\bibliography{aistats}

\end{document}